\documentclass{article}

\usepackage[final]{neurips_2020}
\usepackage[utf8]{inputenc} 
\usepackage[T1]{fontenc}    
\usepackage{hyperref}       
\usepackage{url}            
\usepackage{booktabs}       
\usepackage{amsfonts}       
\usepackage{nicefrac}       
\usepackage{microtype}      
\usepackage{graphicx}
\graphicspath{ {./fig/} }

\title{Hybrid Evolutionary Optimization Approach for Oilfield Well Control Optimization}

\author{%
  Ajitabh Kumar\\
  Visage Technology\\
}

\begin{document}

\maketitle

\begin{abstract}
Oilfield production optimization is challenging due to subsurface model complexity and associated non-linearity, large number of control parameters, large number of production scenarios, and subsurface uncertainties. Optimization involves time-consuming reservoir simulation studies to compare different production scenarios and settings. This paper presents efficacy of two hybrid evolutionary optimization approaches for well control optimization of a waterflooding operation, and demonstrates their application using Olympus benchmark. A simpler, weighted sum of cumulative fluid (WCF) is used as objective function first, which is then replaced by net present value (NPV) of discounted cash-flow for comparison. Two popular evolutionary optimization algorithms, genetic algorithm (GA) and particle swarm optimization (PSO), are first used in standalone mode to solve well control optimization problem. Next, both GA and PSO methods are used with another popular optimization algorithm, covariance matrix adaptation – evolution strategy (CMA-ES), in hybrid mode. Hybrid optimization run is made by transferring the resulting population from one algorithm to the next as its starting population for further improvement. Approximately four thousand simulation runs are needed for standalone GA and PSO methods to converge, while six thousand runs are needed in case of two hybrid optimization modes (GA-CMA-ES and PSO-CMA-ES). To reduce turn-around time, commercial cloud computing is used and simulation workload is distributed using parallel programming. GA and PSO algorithms have a good balance between exploratory and exploitative properties, thus are able identify regions of interest. CMA-ES algorithm is able to further refine the solution using its excellent exploitative properties. Thus, GA or PSO with CMA-ES in hybrid mode yields better optimization result as compared to standalone GA or PSO algorithms.
\end{abstract}

\section{Introduction}
Oilfield planning and operations related decisions such as well placement and production settings have direct impact on cost and revenue for any given project. Hence, field development planning is done to maximize fluid recovery or net present value of cash flow, and is often accomplished using reservoir simulation studies. Simulation results are used to compare the performance of various well configuration and controls. Optimization algorithms are used in a workflow to search for well settings which maximize return on investment [1]. Search involves running thousands of reservoir simulation run for a range of possible solutions and identifying a few solutions of interest for further deliberations.

Optimization algorithms used for such automated workflow are mainly divided in two categories: point based and population based. In this work, population based evolutionary optimization approach is used for oilfield well control optimization problem as these algorithms are relatively simple to program and more effective for a large scale, highly non-linear, and multi-modal problems. Moreover, these algorithms do not require gradient calculations and can also be used for discrete problems. Two popular evolutionary optimization algorithms, genetic algorithm (GA) and particle swarm algorithm (PSO), are used in standalone mode for solving oil well control optimization problem. Next two optimization methods, GA and PSO, are used with another popular optimization method, covariance matrix adaptation – evolution strategy (CMA-ES), in hybrid mode. CMA-ES is also a gradient-free evolutionary optimization algorithm and has been successfully used in different problem settings [6].

Computational requirement of such oilfield well design and control study is quite high, and hence commercial cloud computing is used to handle workload. Parallel programming is used to distribute computational workload among nodes of a large cluster. Two different computational setup are used, one in which a single cluster with a large number of cores is used, and another in which multiple mid-sized clusters are used with a common file server. 

The main contributions of this work are:

\begin{itemize}
  \item Hybrid evolutionary algorithms, GA-CMA-ES and PSO-CMA-ES, are proposed for oilfield well control optimization problem.
  \item Hybrid algorithms are tested using well control optimization study of large Olympus benchmark case.
  \item Robust optimization is done using an ensemble of 10 simulation model realizations, and results are compared with that from a single realization case.
  \item Parallel distribution of computational workload on cloud is done in two ways, one between nodes of large cluster and another between nodes of multiple mid-sized clusters with a common file server.
\end{itemize}

\section{Related Works}
Udy {\it et al.} summarized various approaches for oilfield production optimization and compared algorithms including adjoint methods, particle swarm optimization, simulated annealing and genetic algorithms [1]. Tanaka {\it et al.} compared three different optimization approaches GA, PSO and ensemble-optimization (EnOpt) for Olympus benchmark case [3]. Harb {\it et al.} used hybrid evolutionary optimization method, black hole particle swarm optimization (BHPSO), for simultaneously optimizing well count, location, type, and trajectory [4]. Fonesca {\it et al.} used a hybrid of ensemble optimization EnOpt and gradient free covariance matrix adaptation (CMA) method for well control problem [5], while Bouzarkouna {\it et al.} used CMA-ES method for optimal well placement under uncertainty [2].

\section{Approach}
Olympus benchmark case is a synthetic simulation model based on a typical North Sea offshore oilfield, and is used in this work to compare the effectiveness of different optimization algorithms [11]. Field is a 9km by 3km by 50m in size with minor faults and limited aquifer support. Subsurface uncertainty is captured by 50 realizations with different permeability, porosity, net-to-gross and initial water saturation, of which only 10 random realizations are used in this work to reduce computational requirement. GA and PSO optimization algorithms are first tested for ensemble of 10 realizations, and then going forward a representative realization is used for the remaining studies. A realization in such studies stands for a random simulation model, while ensemble of realizations is often used to capture subsurface uncertainty.

\begin{figure}[t]
  \centering
  \includegraphics[width=\textwidth]{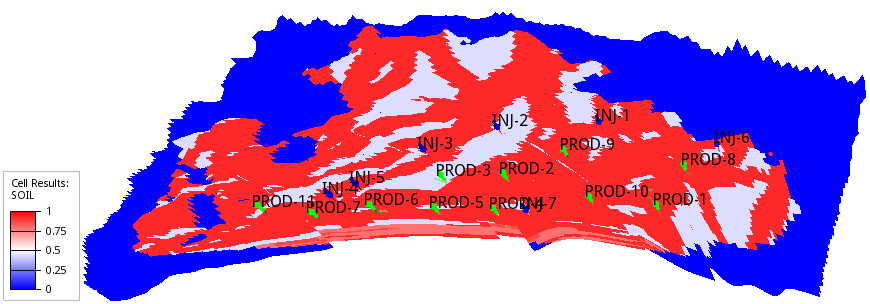}
  \caption{Olympus simulation model overview: initial oil saturation and well locations}
  \label{olympus_model}
\end{figure}

Simulation model has 11 production wells and 7 injection wells which are all on bottom-hole pressure (BHP) control (Figure~\ref{olympus_model}). For the optimization study, each well’s BHP control setting is changed independently at four different pre-defined times during the 20 year simulation period. Thus, there are total 72 different control variables defined as continuous real parameter which can take any value within a given range.

Two different objective functions are used, one as the weighted sum of cumulative fluid production/injection while second as the net present value of discounted cash-flow [1, 14]. This is done to have a rigorous analysis by encoding varying degree of non-linearity among the two objective functions. Weighted sum of cumulative fluid (WCF) is defined as:
\begin{equation}
WCF = Q_{op} - 0.1 * (Q_{wp}+Q_{wi})
\end{equation}
where $WCF$ is weighted sum, $Q_{op}$ is total oil produced, $Q_{wp}$ is total water produced, and $Q_{wi}$ is total water injected. Net present value (NPV) of discounted cash-flow incorporates additional degree of non-linearity and is defined as:
\begin{equation}
NPV = \sum_{f=1}^{N_f}\sum_{\tau=1}^{N_\tau}\frac{Q_{f,\tau}C_f}{(1+d_f)^{\tau-1}}
\end{equation}
where $N_{f}$ is the number of fluid component, $N_{\tau}$ is number of time intervals, $Q_{f,\tau}$ is the volume component produced/injected in time interval $\tau$, $C_{f}$ is the price/cost while $d_{f}$ is discount factor for component $f$.

Parallelization is easy for evolutionary optimization algorithms as objective function calculation for population samples can be done in parallel [3]. This calculation includes making the reservoir simulation run for the population sample (or simulation case), and then determining the required value. Message-Passing Interface (MPI) is used in this study to distribute workload among nodes and thus speed-up calculations. For a large cluster setting, master node assigns simulation cases to different worker nodes of cluster and then accumulates simulation results. A number of small clusters can also be used to parallalize workload if a large cluster is not available or is not efficient. For such a setting, master node on main cluster uses a common file server accessible by smaller clusters which in turn make simulation runs.

\section{Algorithm}
Three different gradient-free, evolutionary optimization algorithms used in this study are genetic algorithm (GA), particle swarm optimization (PSO), and covariance matrix adaptation-evolution strategy (CMA-ES).

Genetic algorithm (GA) has been a popular optimization algorithm and uses three basic steps for population refinement, namely selection, crossover and mutation [9]. For a real-coded genetic algorithm (RCGA) used in this study, real variables themselves are used for population refinement and objective function calculation, as compared to binary encoding done for binary-coded genetic algorithm (BCGA) [10]. In this work, RCGA is used to obviate the need of encoding and decoding control variables. Crossover and mutation steps of GA help the algorithm in exploring complex and multi-modal solution space effectively.

Particle swarm optimization (PSO) uses collective intelligence based on particle position and velocity. It is the exchange of information among particles that leads to the emergence of global patterns [7, 8, 13]. Sample update is done using inertia, personal-best, and global-best values along with random components. It is essential to properly tune hyperparameters for both GA and PSO algorithms to find good solutions and help optimization run converge. Based on domain expertise, the algorithms can be made more exploratory or more exploitative by tuning hyperparameters. The hyperparameters can also be varied during an optimization run to have more exploration in the beginning and more exploitation towards the end of the optimization run.

Covariance matrix adaptation - evolution strategy (CMA-ES), on the other hand, is a stochastic population-based optimization approach where population is updated using weighted mean of samples from the previous generation and a random draw from a multivariate normal distribution [5, 6]. Next, parameters of multivariate normal distribution are updated using objective function evaluations. CMA-ES attempts to learn a second-order model of the underlying objective function and follows a natural gradient approximation of the expected fitness, similar to the approximation of the inverse Hessian matrix in Quasi-Newton methods [6, 18]. For multi-modal functions, where local optima cannot be interpreted as perturbations to an underlying convex (unimodal) topology, its performance can strongly decrease.
\begin{table}[t]
  \caption{Solving Rastrigin test function using GA and PSO}
  \label{rastrigin}
  \centering
  \begin{tabular}{ccccc}
    \toprule
    Dimensions    & Population size  & Iterations & Best cost (GA) & Best cost (PSO)\\
    \midrule
    2     & 40    & 100   & 0.99   & 7.7e-10      \\
    50    & 40    & 100   & 172.1  & 322.0     \\
    50    & 100   & 200   & 97.8   & 158.7      \\
    \bottomrule
  \end{tabular}
\end{table}
\begin{figure}[b]
  \centering
  \includegraphics[scale=0.75]{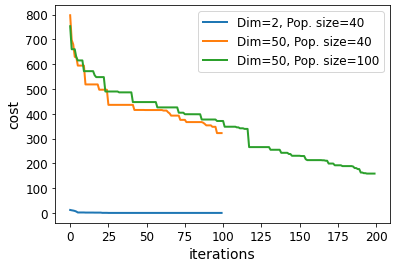}
  \caption{Cost history for solving Rastrigin test function using PSO algirithm}
  \label{pso_rastrigin}
\end{figure}

In continuous domains, the volume of search space grows exponentially with dimensions, and is known as the curse of dimensionality. This phenomenon causes a population size that is capable of adequate search space coverage in low dimensional search spaces to degrade into sparse coverage in high dimensional search spaces. GA and PSO algorithms work well for a problem with smaller number of control variables (or dimensions), but have problem converging when the number of control variables increase [12, 17]. To demonstrate this, both GA and PSO are used for solving Rastrigin function with varied number of dimensions. Rastrigin function has global minimum at zeros for any given number of dimensions and the associated cost is also zero. Cost is calculated as:
\begin{equation}
f = 10d + \sum_{i=1}^{d}[x_i^2-10\cos (2\pi x_i)]
\end{equation}

where $d$ is the number of dimensions, $x_i$ is the value along dimension $i$, and $f$ is the cost. As the dimensionality increases, both GA and PSO require larger population size and iterations, but still do not manage to converge to the global minimum (Table~\ref{rastrigin}, Figure~\ref{pso_rastrigin}). While CMA-ES is not very effective for multi-modal problems, it has been found to have excellent convergence properties for unimodal problems. Thus once GA and PSO algorithms have established the region of interest, CMA-ES can be used to achieve convergence or further improvement. In general, effectiveness of any optimization algorithm depends on the problem class under consideration, and is commonly known as the no free lunch theorem for optimization [19].

In this work, GA and PSO are each first used with ensemble of 10 realizations as well as a single representative realization to establish their efficacy with such optimization problems. Going forward, optimization studies are done only for single representative model in order to cut down on computational cost. For using two optimization methods in a hybrid mode, one optimization algorithm is first allowed to run for a number of generations, and then second algorithm takes over. Two hybrid approaches, GA-CMA-ES and PSO-CMA-ES, are tested and compared with standalone GA and PSO algorithms [14, 15, 16, 18].

\section{Results}
\begin{figure}[b]
  \centering
  \begin{minipage}[b]{0.48\textwidth}
    \includegraphics[width=\textwidth]{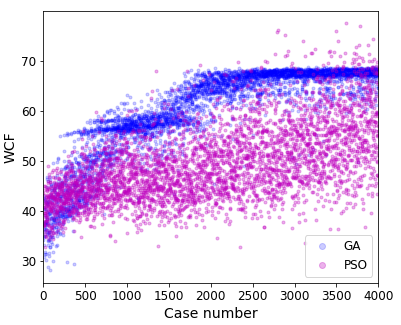}
    \caption{GA and PSO based ensemble optimization run with WCF objective}
    \label{ol_wcf}
  \end{minipage}
  \hfill
  \begin{minipage}[b]{0.48\textwidth}
    \includegraphics[width=\textwidth]{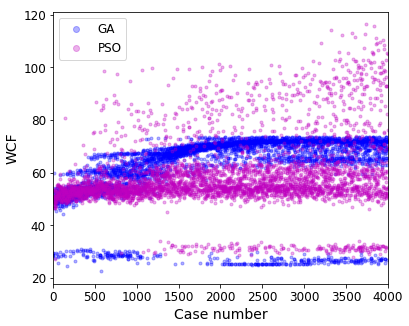}
    \caption{GA and PSO based single realization optimization run with WCF objective}
    \label{ol2_wcf}
  \end{minipage}
\end{figure}

Olympus model has 11 producer wells and 7 injector wells, all on bottom-hole pressure (BHP) constraint (Figure~\ref{olympus_model}). Constraints can be changed independently for each well at four predefined times giving a total of 72 control variables. Well control settings for an ensemble of 10 realizations as well as a single representative realization are first optimized using GA, and then steps are repeated using PSO. Table~\ref{optim_param} shows related parameter values for the optimization runs. A discount factor of 0.08 was used for all fluid components in NPV calculation. For ensemble case, all 10 realizations (or simulation models) are simulated for any given population sample and results are averaged. As can be seen in Figures~\ref{ol_wcf} and~\ref{ol2_wcf}, both GA and PSO algorithms are able to optimize well control settings and yield higher weighted cumulative fluid as compared to the default case. As sampling is randomized in both the algorithms, every optimization run will proceed in unique fashion. Thus multiple optimization runs are required for comparing different algorithms. Going forward, only single representative reservoir model is optimized to cut down on computational requirement.
\begin{table}[t]
  \caption{Optimization study parameters}
  \label{optim_param}
  \centering
  \begin{tabular}{lr||lr}
    \toprule
    Population size & 40 & Oil price & \$ 40 per barrel \\
    Iterations (GA or PSO) & 100 & Produced water cost & \$ 4 per barrel \\
    Iterations (Hybrid) & 150 & Injected water cost & \$ 2 per barrel \\
    \bottomrule
  \end{tabular}
\end{table}

\begin{figure}[t]
  \centering
  \includegraphics[scale=0.65]{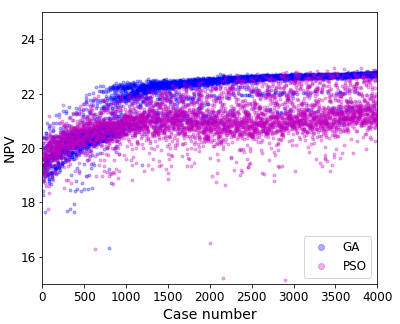}
  \caption{GA and PSO based single realization optimization run with NPV objective}
  \label{ol2_npv}
\end{figure}

To further validate algorithm efficacy, objective function is next changed to net present value (NPV) of discounted cash-flow, and the results for GA and PSO based optimization runs are shown in Figure~\ref{ol2_npv}. Again algorithms seem to be effective in handling such simulation based oil production optimization problem. As compared to weighted sum of cumulative fluid (WCF) based optimization, NPV based optimization yields muted increase over default/base case because of the discounting.

\begin{figure}[b]
  \centering
  \begin{minipage}[b]{0.48\textwidth}
    \includegraphics[width=\textwidth]{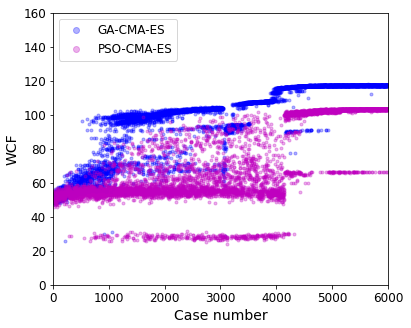}
    \caption{Hybrid GA-CMA-ES and PSO-CMA-ES based single realization optimization run with WCF objective}
    \label{ol2_hybrid_wcf}
  \end{minipage}
  \hfill
  \begin{minipage}[b]{0.48\textwidth}
    \includegraphics[width=\textwidth]{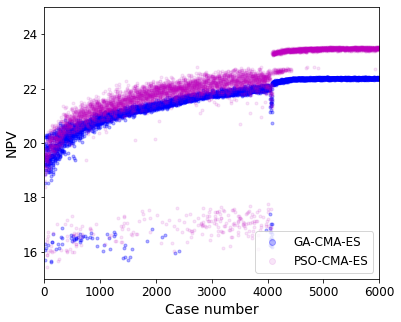}
    \caption{Hybrid GA-CMA-ES and PSO-CMA-ES based single realization optimization run with NPV objective}
    \label{ol2_hybrid_npv}
  \end{minipage}
\end{figure}

Hybrid optimization run is made by transferring the resulting population from one algorithm to the next as its starting population for further improvement. Total 6000 simulation runs are made for such studies, out of which first 4000 runs (or 100 generations) are done using primary algorithm followed by 2000 runs (or 50 generations) using secondary algorithm. Two different hybrid optimization settings are: GA followed by CMA-ES (GA-CMA-ES) and PSO followed by CMA-ES (PSO-CMA-ES). Once region of interest is found using GA or PSO, CMA-ES algorithm further refines the solution as it is a better algorithm for local exploitation. Figures~\ref{ol2_hybrid_wcf} and~\ref{ol2_hybrid_npv} show that CMA-ES is able to further improve upon the solution, and thus GA-CMA-ES and PSO-CMA-ES approach of hybrid optimization is better as compared to standalone GA or PSO. There is a marked jump in objective function value after 4000 runs when CMA-ES takes over from GA or PSO. It needs to be mentioned that CMA-ES does not have good exploratory mechanism for multi-modal solution space, and hence it does not perform as good as GA or PSO in standalone mode. Domain expertise is crucial for success of such optimization studies as their design would vary depending on underlying solution space and objective function.

\begin{figure}[t]
  \centering
  \includegraphics[width=\textwidth]{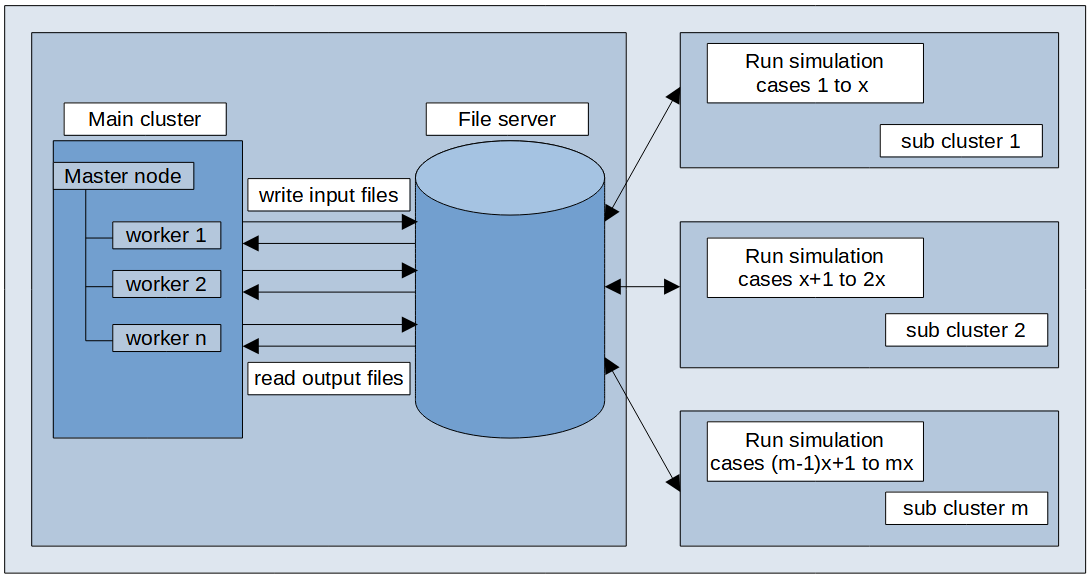}
  \caption{Compute resource setup with multiple clusters and common file-server}
  \label{config}
\end{figure}

Population based algorithms can be easily programmed in parallel mode with objective function for each sample being calculated using a separate node. In the calculation of objective function for a given population sample, compute intensive step is making the reservoir simulation run. Either a large cluster can be used to divide workload among cores, or multiple mid-size clusters can be used with a common file server. By reading from and writing to a common file server (Figure~\ref{config}), multiple clusters can be used to carry-out compute intensive simulation runs, thus reducing turn-around time. 

\section{Conclusion}
Evolutionary algorithms are efficient at solving oilfield well control optimization problem. Two such algorithms are used in hybrid mode to explore multi-modal solution space and attain convergence. GA-CMA-ES and PSO-CMA-ES hybrid optimization approach perform better as compared to using GA or PSO for such optimization problem. For the hybrid optimization, GA or PSO algorithm is first used to explore solution space and identify region of interest, and then CMA-ES is used to further refine the solution. Domain expertise is crucial for success of such optimization studies as their design would vary depending on solution space and objective function. Cloud computing and parallel programming are used to handle the computational workload. Two different cluster settings, one with single large cluster and other with multiple mid-sized clusters, have been used to handle high computational workload.

\begin{ack}
Research was made possible by generous cloud credits provided by Amazon Web Services, DigitalOcean, Google Cloud and IBM Cloud. Olympus benchmark reservoir model was made available by the Netherlands Organization for Applied Scientific Research (TNO).
\end{ack}

\section*{References}

\medskip

\small

[1] Udy, J., Hansen, B., Maddux, S., Petersen, D., Heilner, S., Stevens, K., Lignell, D.\ \& Hedengren, J.D.\ (2017) Review of field development optimization of waterflooding, EOR, and well placement focusing on history matching and optimization algorithms. {\it Processes} {\bf 5}(34).

[2] Bouzarkouna, Z., Ding, D.Y.\ \& Auger, A.\ (2012) Well placement optimization under uncertainty with CMA-ES using the neighborhood. {\it $13^{th}$ European Conference on the Mathematics of Oil Recovery}.

[3] Tanaka, S., Wang, Z., Dehghani, K., He, J., Velusamy, B.\ \& Wen, X.\ (2018) Large scale field development optimization using high performance parallel simulation and cloud computing technology. {\it SPE Annual Technical Conference and Exhibition}.

[4] Harb, A., Kassem, H.\ \& Ghorayeb, K.\ (2019) Black hole particle swarm optimization for well placement optimization. {\it Computational Geosciences}, {\bf 24} (6):1979-2000.

[5] Fonseca, R.M., Leeuwenburgh, O., Van den Hof, P.M.J.\ \& Jansen, J.D.\ (2013) Improving the ensemble optimization method through covariance matrix adaptation (CMA-EnOpt). {\it SPE Reservoir Simulation Symposium}.

[6] Hansen, N.\ (2016) The CMA evolution strategy: a tutorial. arXiv preprint \href{https://arxiv.org/abs/1604.00772}{arXiv:1604.00772}.

[7] Wang, D., Tan, D.\ \& Liu, L.\ (2018) Particle swarm optimization algorithm: an overview. {\it Soft Computing}, {\bf 22}:387-408.

[8] Sengupta, S., Basak, S.\ \& Peters, R.A.\ (2019) Particle swarm optimization: a survey of historical and recent developments with hybridization perspectives. {\it Machine Learning and Knowledge Extraction}, {\bf 1}:157-191.

[9] Katoch, S., Chauhan, S.S.\ \& Kumar, V.\ (2021) A review on genetic algorithm: past, present, and future. {\it Multimedia Tools and Applications}, {\bf 80}:8091-8126.

[10] Chuang, Y.C., Chen, C.T.\ \& Hwang, C.\ (2015) A real-coded genetic algorithm with a direction-based crossover operator. {\it Information Sciences}, {\bf 305}:320-348.

[11] Fonseca, R.M., Rossa, E.D., Emerick, A.A., Hanea, R.G.\ \& Jansen, J.D.\ (2020) Introduction to the special issue: overview of OLYMPUS optimization benchmark challenge. {\it Computational Geosciences}, {\bf 24} (6):1933-1941. 

[12] Chen, S., Montgomery, J.\ \& Bolufé-Röhler, A.\ (2015) Measuring the curse of dimensionality and its effects on particle swarm optimization and differential evolution. {\it Appl Intell}, {\bf 42}:514-526.

[13] Zhang, Y., Wang, S.\ \& Ji, G.\ (2015) A comprehensive survey on particle swarm optimization algorithm and its applications. {\it Mathematical Problems in Engineering}, {\bf 2015}.

[14] Baumann, E.J.M, Dale, S.I.\ \& Bellout, M.C.\ (2020) FieldOpt: A powerful and effective programming framework tailored for field development optimization. {\it Computers and Geosciences}, {\bf 135}.

[15] Xu, P., Luo, W., Lin, X., Qiao, Y.\ \& Zhu, T.\ (2019) Hybrid of PSO and CMA-ES for global optimization. {\it IEEE Congress on Evolutionary Computation}.

[16] Santos, J., Ferreira, A.\ \& Flintsch, G.\ (2019) An adaptive hybrid genetic algorithm for pavement management. {\it International Journal of Pavement Engineering}, {\bf 20} (3):266-286.

[17] Clerc, M.\ \& Kennedy, J.\ (2002) The particle swarm - explosion, stability, and convergence in a multidimensional complex space. {\it IEEE Transactions on Evolutionary Computation}, {\bf 6} (1):58-73.

[18] Muller, C.L., Baumgartner, B.\ \& Sbalzarini, I.F.\ (2009) Particle swarm CMA evolution strategy for the optimization of multi-funnel landscapes. {\it IEEE Congress on Evolutionary Computation}, pp. 2685-2692.

[19] Wolpert, D.H.,\ \& Macready, W.G.\ (1997) No free lunch theorems for optimization. {\it IEEE Transactions on Evolutionary Computation}, {\bf 1} (1):67-82.

\end{document}